\documentclass[11pt]{article}
\usepackage[final]{acl}
\usepackage{times}
\usepackage{latexsym}
\usepackage[T1]{fontenc}
\usepackage[utf8]{inputenc}
\usepackage{microtype}
\usepackage{inconsolata}
\usepackage{graphicx}
\usepackage{booktabs}
\usepackage{amsmath}
\usepackage{amssymb}
\usepackage{xcolor}
\usepackage{multirow}
\usepackage{url}

\newcommand{\fig}[2]{\IfFileExists{figures/#1.pdf}{\includegraphics[width=#2]{figures/#1}}{\fbox{\parbox[c][3cm][c]{#2}{\centering\small\texttt{\detokenize{figures/#1.pdf}}}}}}

\newcommand{\Cov}{\operatorname{Cov}}
\newcommand{\Var}{\operatorname{Var}}
\newcommand{\E}{\mathbb{E}}
\newcommand{\dr}{\Delta r}

\title{Forecasting How RLHF Responds to Reward-Model Edits Before Training}
\author{Zixiao Wang, Duzhen Zhang, Xingbo Du, Dongxia Wu, Le Song \\
  Mohamed bin Zayed University of Artificial Intelligence \\
  \texttt{\{zixiao.wang,duzhen.zhang,xingbo.du,dongxia.wu,le.song\}@mbzuai.ac.ae}}

\hypersetup{pdftitle={Forecasting How RLHF Responds to Reward-Model Edits Before Training},pdfauthor={Zixiao Wang, Duzhen Zhang, Xingbo Du, Dongxia Wu, Le Song}}

\begin{document}
\maketitle

\begin{abstract}
Responses with headers, lists and greater length receive high scores from reward models (RMs), and RLHF policies learn to produce these features even when true quality does not improve. Practitioners therefore edit their RMs, but they learn whether an edit works only after an RL run. We show that the response of a policy to an RM edit can be forecast before training, from samples of the current policy and the scores of the old and new RM. If we read the RM as a set of prices, the first-order theory of KL-regularized RL implies that each feature moves in proportion to its within-prompt covariance with the change in reward. We implant controlled price changes into retrained RMs under a $2^{4-1}$ fractional factorial design with a self-distilled placebo, run 54 RL runs from six parent policies, and fix every forecast before the runs. With the covariance predictor, we predict the direction of all 20 changes in supply whose forecast has the same sign at every parent and explain 82\% of the standardized change. With counterfactual prices, the classical alternative, we explain 50\%. The forecast also holds under GRPO and for real RM upgrades, where we explain 66\% out of sample with a coefficient frozen from the factorial runs. For rent, the gap between the proxy reward and a gold reward from stronger judges, we rank RM edits correctly with the same covariance over the first 30 steps (Spearman 0.94). This measures the horizon of the forecast: by step 200, the ranking of the factorial edits reverses under both RLOO and GRPO, while the real upgrades keep their order. When we vary the KL coefficient $\beta$, the initial response does not depend on $\beta$, and the response saturates below the KL equilibrium. In one experiment, the long-run response to a price rise is consistent with an additional implicit regularizer $\beta_0 \approx 0.003$--$0.007$.
\end{abstract}

\section{Introduction}\label{sec:intro}

In reinforcement learning from human feedback (RLHF), a policy is trained against a learned reward model \citep{christiano2017deep, ziegler2019fine, stiennon2020learning, ouyang2022training}. When the optimization goes far enough, proxy reward keeps rising while true quality stalls or falls \citep{gao2023scaling, pan2022effects}. Much of this gap comes from features that receive high scores from the reward model but have no value for a careful reader, such as length \citep{singhal2024long, chen2024odin} and formatting \citep{zhang2025lists}. \citet{rentseek2026} call the gap \emph{rent} and show that the rent of a policy follows the prices of its reward model.

The usual response is to edit the reward model. Teams add penalties, retrain on more or cleaner data, combine several models in an ensemble \citep{coste2024reward, rame2024warm}, or audit a fraction of samples against a stronger judge \citep{rentseek2026}. Each edit is a bet. Only after an RL run do teams learn whether rent fell and what the policy produces instead \citep{baker2025monitoring}. A run costs hundreds of GPU-hours, so teams evaluate most edits on short runs or not at all.

We ask whether the effect of an edit can be computed before RL starts. We take the economic reading literally (Appendix~\ref{app:glossary} defines each term). In this reading, each feature of a response has a price under the reward model, and a policy under KL-regularized RL supplies features in response to these prices. Two classical results then apply: the closed-form optimum of KL-regularized RL \citep{korbak2022rl, rafailov2023direct} and the first-order dynamics of policy updates \citep{price1970selection}. Under both, the change in a feature is proportional to $\Cov(f_k, \dr)$, its within-prompt covariance with the change in reward. We call this quantity the \emph{covariance predictor}. From producer theory, we obtain a second quantity: the change in the counterfactual prices of features, multiplied by their covariance \citep{hotelling1932edgeworth}. We call it the \emph{price predictor}. The two agree only if the reward model values features additively.

We test both predictors with price changes that we control. We implant prices with soft labels into retrained copies of the proxy, the reward model that the policy optimizes, so that each treated model differs from a self-distilled placebo only in the prices of chosen features. We vary four features at once in a $2^{4-1}$ fractional factorial design \citep{box1961fractional}. From six parent policies of the \citet{rentseek2026} testbed, we run 54 RL runs, and we compute every forecast and fix the analysis plan before the first run, following the logic of preregistration \citep{nosek2018preregistration}. We then apply the same forecast, with no refitting, to real upgrades of the reward model and fix it before the runs as well.

\begin{figure}[t]
\centering
\fig{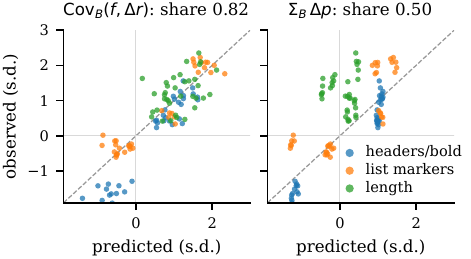}{\columnwidth}
\caption{Predicted against observed change in supply after 200 steps, relative to the placebo RM, for 6 factorial arms $\times$ 6 parents and the three primary features. Units are within-prompt standard deviations at the branch point, and each predictor has one fitted coefficient per feature. Left: covariance predictor $\Cov_B(f, \dr)$. Right: price predictor $\Sigma_B\,\Delta p$. We computed both predictors at the branch point before any RL step.}
\label{fig:headline}
\end{figure}

Our contributions are as follows.
\begin{itemize}
\setlength\itemsep{0pt}
\item \textbf{A forecast of supply that works before training.} With the covariance predictor, we predict the direction of all 20 stable changes in supply and explain 82\% of the standardized response after 200 steps. With the price predictor, we explain 50\% (Figure~\ref{fig:headline}). The gap lies in length, which no list of priced features covers. On three real upgrades of the reward model, we explain 66\% out of sample with a frozen coefficient, and it holds under GRPO.
\item \textbf{A forecast of rent, with a measured horizon.} With rent in place of a feature, the same covariance agrees with the ranking of edits by their effects on rent and gold reward over the first 30 steps (Spearman 0.94 and 0.89). We also recover the ranking of 13 reward patches from prior work, such as audits and length penalties (0.97). By step 200, the ranking of our factorial arms reverses under RLOO and GRPO, while real upgrades keep their order. In exploratory analyses, the reversal follows the extra distance that the policy travels, and when we recompute the covariance at a checkpoint, we see the incentive after a price cut grow or change sign.
\item \textbf{Evidence that RL stops short of KL equilibrium.} When we vary $\beta$ and time jointly, the initial response does not depend on $\beta$, and the response saturates at about half the level that $1/\beta$ implies. After a price rise, the asymptote and the time constant fit an effective coefficient $\beta + \beta_0$ ($\beta_0 \approx 0.0065$; 0.0029 in a prospective test).
\end{itemize}

\section{Related Work}\label{sec:related}

\paragraph{Reward overoptimization.} Proxy reward diverges from true reward under optimization \citep{gao2023scaling, skalse2022defining}, and the divergence grows with the capability of the optimizer \citep{pan2022effects}. \citet{manheim2018categorizing} sort such failures into variants of Goodhart's law, and \citet{casper2023open} list them among the open problems of RLHF. They persist under iterated RLHF \citep{wolf2025reward}. We study a narrower question: given a specific edit to the reward model, in which direction and by how much will a policy move?

\paragraph{Biases of reward models and their repair.} Longer responses \citep{singhal2024long, dubois2024length} and heavier formatting \citep{zhang2025lists} receive higher scores from reward models, and reward models inherit value biases from pretraining \citep{christian2026reward}. Proposed repairs include separate heads for length \citep{chen2024odin}, adaptive length debiasing \citep{bu2025beyond}, causal rewards \citep{wang2025beyond}, data augmentation \citep{liu2025rrm}, uncertainty penalties \citep{zhang2024overcoming}, ensembles \citep{coste2024reward, eisenstein2024helping} and weight averaging \citep{rame2024warm}. A policy that is penalized for a detected hack can learn to hide it \citep{baker2025monitoring}. In these works, a repair is evaluated after training. Our forecast applies to any of them before training, because we need only the scores of the repaired model.

\paragraph{Evaluating reward models.} Benchmarks such as RewardBench \citep{lambert2025rewardbench} and RM-Bench \citep{liu2025rmbench} measure the accuracy of a reward model on preference pairs, and RM-Bench includes pairs that test whether a model prefers style, such as length and markdown, over substance. \citet{frick2025evaluate} relate such proxy metrics to the outcome of RLHF and find that accuracy on human preferences predicts the outcome best. On these benchmarks, a reward model receives one score. We ask a different question: after an edit to the reward model, in which direction and by how much will a given policy change each behavior? We answer it from the difference between the two models on samples of that policy.

\paragraph{Rent and the economics of alignment.} \citet{rentseek2026} measure the gap between proxy and gold reward as rent, a notion from the economics of regulation \citep{tullock1967welfare, krueger1974political}. Alignment has also been framed as incomplete contracting \citep{hadfield2019incomplete} and as an equilibrium under finite evaluation \citep{wang2026equilibrium}. We use a third tool from economics, the theory of supply. In hedonic price models \citep{rosen1974hedonic}, the value of a product is a sum over its characteristics, and in producer theory, supply follows from these values \citep{hotelling1932edgeworth}. We check whether a policy trained by RL behaves like such a producer.

\paragraph{Predicting the effect of training.} With influence functions \citep{koh2017understanding} and datamodels \citep{ilyas2022datamodels}, one can predict how the outputs of a model change after a change in its training data. \citet{ren2025learning} decompose the dynamics of fine-tuning step by step, and the optimum of KL-regularized RL is an exponential tilt of the reference policy \citep{korbak2022rl, rafailov2023direct, beirami2025theoretical}. The covariance form of the response appears in producer theory \citep{hotelling1932edgeworth}, statistical physics \citep{kubo1966fluctuation} and evolutionary biology \citep{price1970selection}. We turn this form into forecasts for edits to a reward model and test them under controlled price changes.

\paragraph{Poisoning reward models.} When preferences are planted in the training data of a reward model, the trained policy changes \citep{rando2024universal, wang2024rlhfpoison}. We use planting as a measurement instrument, with several doses, several features at once and a placebo, so that we can measure the response to a known change in price.

\section{A Price Theory of Feature Supply}\label{sec:theory}

We define prices on features, derive two readings of how a policy supplies them, and arrive at the two predictors that we test. Throughout, the change is local: we take a checkpoint of RL, the \emph{branch point}, edit the reward model there, and continue RL for a few hundred steps. A subscript $B$ marks a quantity that we compute on samples at the branch point.

A prompt $x$ elicits a response $y \sim \pi(\cdot\mid x)$ with features $f(y) = (f_1, \dots, f_K)$. A feature is a count, such as the number of headers, or an indicator, such as the presence of a closing summary. As in \citet{rentseek2026}, we standardize every reward model with fixed constants. Let $z_p$ be the proxy reward and $z_g$ the mean of two gold rewards. The rent of a response is $\rho = z_p - z_g$. The \emph{supply} of feature $k$ is its mean $s_k = \E_{x,\,y\sim\pi}[f_k]$.

The counterfactual price of feature $k$ under reward model $m$ is the mean drop in score when we delete $k$ from a response and keep every other character:
\begin{equation}
p_k^m = \E\left[z_m(x,y) - z_m(x,y^{-k})\right].
\label{eq:price}
\end{equation}
We measure prices per pair, that is, per deletion of the whole feature. We obtain a price per unit when we divide by the mean number of units that the deletion removes.

RL with reward $r$ and a KL penalty of weight $\beta$ toward a reference policy $\pi_0$ has the optimum $\pi^*(y\mid x) \propto \pi_0(y\mid x)\exp(r(x,y)/\beta)$ \citep{korbak2022rl, rafailov2023direct}. Near the current point, we write the reward as $r = a(x) + \sum_k p_k f_k + e$ and define the profit function $\Phi(p) = \beta\log\E_{\pi_0}[\exp(r/\beta)]$. Then
\begin{equation}
\begin{gathered}
\frac{\partial \Phi}{\partial p_k} = \E_{\pi^*}[f_k], \\
S_{kj} \equiv \frac{\partial \E_{\pi^*}[f_k]}{\partial p_j} = \frac{\Cov(f_k, f_j\mid x)}{\beta}.
\end{gathered}
\label{eq:hotelling}
\end{equation}
Supply is the gradient of profit, as in Hotelling's lemma \citep{hotelling1932edgeworth}. The supply matrix is the covariance divided by $\beta$, as in the fluctuation--response relation of a Gibbs distribution \citep{kubo1966fluctuation}. We take covariances within prompt and average them over prompts, because the tilt applies to each prompt separately.

A policy need not reach $\pi^*$ for a law of this form to hold. After one exponential-weights step $\pi_{t+1} \propto \pi_t\exp(\eta A)$ with advantage $A$, the mean of a feature changes by $\Delta\E[f_k] \approx \eta\,\Cov_{\pi_t}(f_k, r)$, the selection term of the Price equation \citep{price1970selection}. Here the coefficient grows with the learning rate and the number of steps, and it does not depend on $\beta$.

Under both readings, the response of supply to an edit $\dr(y) = r_{\text{new}}(y) - r_{\text{old}}(y)$, with the standardization used in RL, is
\begin{equation}
\Delta s_k \approx \kappa\,\E_x\left[\Cov_x(f_k, \dr)\right],
\label{eq:covdr}
\end{equation}
where $\kappa$ is $1/\beta$ at equilibrium and about $\eta$ times the number of steps early in training. We need only samples from the policy and scores from the two reward models to evaluate Eq.~\ref{eq:covdr}, and we need no list of priced features. If the edit is a linear price change on the measured features, so that $\dr = \sum_j \Delta p_j f_j$, Eq.~\ref{eq:covdr} reduces to the price predictor
\begin{equation}
\Delta s \approx \kappa\,\Sigma\,\Delta p,
\label{eq:sigdp}
\end{equation}
with $\Sigma$ the within-prompt covariance of the features and $\Delta p$ the change in prices per unit.

\paragraph{Where the two predictors part.} The reduction fails in two ways. First, after an edit, rewards can change outside the span of the priced features. We expect length to be the main case: under a reward model with a higher price for formatting, the longer responses that carry formatting may also receive higher rewards. Such a change is part of $\dr$ but not of $\Delta p$. Second, the reward model may value features non-additively. For goods $a$ and $b$, we identify the interaction $h_{ab}$ in $z = \dots + p_a f_a + p_b f_b + h_{ab} f_a f_b$ by
\begin{equation}
\Delta z_{ab} - (\Delta z_a + \Delta z_b) = h_{ab} f_a f_b ,
\label{eq:interaction}
\end{equation}
where $\Delta z_a$, $\Delta z_b$ and $\Delta z_{ab}$ are the changes in score when we delete $a$, $b$ or both; a change is negative when the score falls.
On 18,537 responses of our parent policies, the proxy loses 0.174 standard deviations when we delete headers and bold, 0.028 when we delete list markers, and 0.485 when we delete both. The joint loss is 2.4 times the sum, so $h_{ab} f_a f_b = -0.485 + 0.174 + 0.028 = -0.283 < 0$. In the valuation of the reward model, the two goods are substitutes, and one more element on a response that already carries a dozen is worth less than the price per unit in Eq.~\ref{eq:sigdp}.

\paragraph{Relaxation.} Along one feature direction, with $\pi_\theta \propto \pi_0\exp(\theta f)$, the natural-gradient flow of the KL-regularized objective is $\dot\theta = \eta(p - \beta\theta)$. After a price shock at $t_0$, the response is
\begin{equation}
\Delta s(t) \approx \frac{1 - e^{-\eta\beta(t-t_0)}}{\beta}\,\Cov\cdot\Delta p .
\label{eq:relax}
\end{equation}
We derive three fingerprints from Eq.~\ref{eq:relax}: the asymptote $A$ scales with $1/\beta$, the time constant $\tau$ scales with $1/\beta$, and the initial slope $A/\tau$ does not depend on $\beta$. If all three hold, RL relaxes toward the tilted optimum. If only the third holds, $\kappa$ is an effective learning rate (Appendix~\ref{app:theory}).

\paragraph{Cross effects and rent.} In Eq.~\ref{eq:hotelling}, the off-diagonal terms correspond to cross effects: when two goods covary positively within prompt, a higher price for one raises the supply of the other, so they are complements in supply. If we replace $f_k$ in Eq.~\ref{eq:covdr} by rent $\rho$ or by gold reward $z_g$, we obtain a forecast of the change in rent, $\Delta\rho \approx \kappa\,\E_x[\Cov_x(\rho, \dr)]$, and of the change in gold reward. For this forecast, we do not need to know which features the policy will turn to.

\section{Experimental Design}\label{sec:design}

To test the predictors, we need price changes that we control and a protocol that fixes every forecast in advance.

\paragraph{Testbed.} We use the testbed of \citet{rentseek2026}. The policy is Qwen3-1.7B-Base \citep{yang2025qwen3} after supervised fine-tuning (SFT) on Tulu~3 conversations \citep{lambert2025tulu3}. We train it with RLOO \citep{ahmadian2024back} in verl \citep{sheng2025hybridflow}, with $\beta = 0.005$, 64 prompts and 8 samples per step, and a learning rate of $2\times10^{-6}$. Prompts come from UltraFeedback \citep{cui2024ultrafeedback}. The proxy is a reward model on Qwen3-1.7B-Base, trained on 3,000 preference pairs, and we define gold reward with two Skywork-Reward-V2 models \citep{liu2026skywork}. Our six parents are policies trained for 600 steps without the audits of \citet{rentseek2026}: three seeds of the original runs and three further seeds. Rent at the branch points is 0.68 to 0.71 standard deviations.

\paragraph{Goods.} We price four goods: headers and bold (F1a), list markers (F1b), a closing summary (F2) and a list of external resources (F4). A parent response carries about 16.5 header and bold elements and 15.5 list markers, 39\% of responses close with a summary, and 3\% carry a resource list. We find each good with a rule-based detector and remove it with a deletion rewrite that keeps every other character (Appendix~\ref{app:goods}). Following the pre-specified plan, F1a, F1b and length $L$ in policy tokens form the primary features. F2 and F4 are secondary. We call a feature a good when we can price it by deletion, so length is a feature but not a good.

\paragraph{Soft-label implants.} We train every reward model with the recipe of the proxy: the same base model, seed, learning rate and batch size, and the 3,000 pairs of the proxy in the same order. To these pairs we add 600 pure deletion pairs for each good (597 for F4), which we build from responses of three SFT policies, and 600 content pairs, each of which deletes one random sentence, so that the price of content stays anchored. All reward models train on the same 5,997 pairs, and only the labels differ. Let $\Delta_i$ be the gap in raw score of the proxy between the original and the rewrite of pair $i$. We train with a cross-entropy loss on soft labels. For the \emph{placebo}, we use the label $\sigma(\Delta_i)$ on every pair, a self-distillation \citep{hinton2015distilling} under which the prices of the proxy stay unchanged. For a \emph{treated} model, we use $\sigma(\Delta_i + \delta_k)$ on the pairs of good $k$ and $\sigma(\Delta_i)$ elsewhere. With this shift of the label, the price of good $k$ rises by $\delta_k$ raw reward units. We standardize every reward model with the constants of the proxy (mean 6.639, s.d.\ 2.161), in all comparisons and in RL.

\paragraph{Factorial design and validation.} The eight treated models follow a $2^{4-1}$ fractional factorial design with generator $D = ABC$. The design has resolution IV, and F1a$\times$F1b is aliased with F2$\times$F4. The doses are $\delta = \pm 4$ for F1a and F1b, $+2$ or $-1.5$ for F2, and $0$ or $+1.5$ for F4, whose supply leaves no room below zero. We name each arm by its levels of F1a, F1b, F2 and F4, with 1 for the higher dose. We define the off-target noise as the root mean square movement of treated models on goods that they do not implant, measured on the same responses. A treated model passes if its price change on each implanted primary good has the designed sign and is at least three times this noise. On SFT samples for 1,000 held-out prompts, the threshold is 0.195, and all 16 implanted primary goods pass. On parent responses at the branch points, the threshold is 0.213, and 15 of 16 pass (Table~\ref{tab:implant}; Appendix~\ref{app:implant}). In all analyses, we use measured reward differences and prices rather than designed doses.

\begin{table}[t]
\centering
\small
\setlength{\tabcolsep}{2.5pt}
\resizebox{\columnwidth}{!}{%
\begin{tabular}{lrrrrrrrr}
\toprule
 & \multicolumn{4}{c}{SFT samples} & \multicolumn{4}{c}{Parent responses} \\
\cmidrule(lr){2-5}\cmidrule(lr){6-9}
Arm & F1a & F1b & F2 & F4 & F1a & F1b & F2 & F4 \\
\midrule
0000 & $-$0.69 & $-$0.73 & $-$0.45 & 0.01 & $-$0.67 & $-$0.28 & $-$0.28 & $-$0.02 \\
1001 & 0.83 & $-$0.94 & $-$0.48 & 0.37 & 0.74 & $-$0.20 & $-$0.14 & $-$0.01 \\
0101 & $-$0.72 & 0.75 & $-$0.42 & 0.39 & $-$0.87 & 0.44 & $-$0.46 & 0.18 \\
1100 & 0.65 & 0.84 & $-$0.41 & 0.08 & 0.65 & 0.34 & $-$0.22 & 0.02 \\
0011 & $-$0.79 & $-$0.99 & 0.74 & 0.38 & $-$0.45 & $-$0.38 & 0.35 & 0.39 \\
1010 & 0.82 & $-$0.98 & 0.67 & $-$0.02 & 0.67 & $-$0.22 & 0.21 & 0.02 \\
0110 & $-$0.74 & 0.86 & 0.70 & 0.10 & $-$0.46 & 0.49 & 0.30 & 0.13 \\
1111 & 0.61 & 0.92 & 0.65 & 0.43 & 0.60 & 0.24 & 0.10 & 0.09 \\
\midrule
Threshold & \multicolumn{4}{c}{0.195} & \multicolumn{4}{c}{0.213} \\
\bottomrule
\end{tabular}}
\caption{Implanted price changes relative to the placebo RM, per pair, on the common standardized scale. Parent columns average six branch points. The threshold is three times the off-target noise and applies to the primary goods F1a and F1b.}
\label{tab:implant}
\end{table}

\paragraph{Runs and windows.} We first train each parent for 50 steps against the placebo from a fresh optimizer state. From this branch point, nine arms (the placebo and the eight treated models) run 200 steps each, with the same weights, RL seed and prompt order, a fresh optimizer and $\beta = 0.005$. In total, we run 54 runs. We evaluate supply on 500 held-out prompts every 5 steps, and we compare each arm with the placebo of the same parent in an early window (steps 20--30) and an end window (steps 160--200). In every run, we also score the samples with the original proxy, which no run optimizes, so that we can measure rent on one fixed ruler.

\paragraph{Pre-specification.} Before any run, we sampled 16 responses for each of 200 prompts at every branch point and scored them with all reward models. From these samples we computed both predictors for every arm, parent and feature, and the forecasts of rent and gold reward. In the pre-specified analysis, we standardize every quantity by its within-prompt standard deviation at the branch point. We fit one coefficient $\kappa$ per predictor and one per feature, report the explained share $1 - \lVert \Delta s - \kappa\,\hat{s}\rVert^2 / \lVert\Delta s\rVert^2$, and count a sign only where the prediction has the same sign at every parent. In Appendix~\ref{app:plan}, we list every pre-specified analysis and mark each later analysis as exploratory.

\section{Forecasting Supply}\label{sec:supply}

\begin{table}[t]
\centering
\small
\setlength{\tabcolsep}{3pt}
\begin{tabular}{lcccc}
\toprule
 & \multicolumn{2}{c}{Steps 20--30} & \multicolumn{2}{c}{Steps 160--200} \\
\cmidrule(lr){2-3}\cmidrule(lr){4-5}
Predictor & Sign & Share & Sign & Share \\
\midrule
Covariance & 18/20 & 0.81 & 20/20 & 0.82 \\
Price & 17/20 & 0.58 & 20/20 & 0.50 \\
\midrule
\multicolumn{5}{l}{\emph{Robustness}} \\
Covariance, common $\kappa$ & & 0.69 & & 0.82 \\
Price, common $\kappa$ & & 0.51 & & 0.45 \\
Covariance, five features & & 0.80 & & 0.82 \\
Price, five features & & 0.57 & & 0.51 \\
Covariance, + content price & & 0.87 & & 0.82 \\
Price, + content price & & 0.70 & & 0.73 \\
Covariance, replication & & 0.88 & & 0.66 \\
Price, replication & & 0.67 & & 0.59 \\
\bottomrule
\end{tabular}
\caption{Pre-specified forecasts of supply for 6 arms $\times$ 6 parents. Sign: of the 24 changes (6 arms $\times$ 4 goods), those whose forecast has the same sign at every parent, and how many of them have the observed sign. Share: explained share of the standardized response of F1a, F1b and length, with one coefficient per feature unless noted. In ``+ content price'', the measured change in the price of content enters as a covariate. ``Replication'' is a second batch on three parents in which each RM has its own reward scale (\S\ref{sec:rent}).}
\label{tab:predict}
\end{table}

\paragraph{The covariance predictor forecasts supply.} In two arms, which raise the price of summaries and lower the price of headers, length runs away (48\% and 30\% of responses at the length limit), so under the pre-specified rule we analyze the other six (Appendix~\ref{app:supply}). We list both predictors in Table~\ref{tab:predict}. We call a forecast \emph{stable} when it has the same sign at every parent; 20 of the 24 changes (6 arms $\times$ 4 goods) are stable. With the covariance predictor, we predict the direction of all 20 in the end window and explain 82\% of the standardized response of the primary features. The 95\% interval, by bootstrap over parents, runs from 0.80 to 0.88. With the price predictor, we also predict the 20 directions but explain 50\% (0.46 to 0.53), and the interval of the difference runs from 0.29 to 0.39. The ordering holds in both windows, with a common coefficient or one per feature, and with all five features. It also holds when we add the change in the price of content as a covariate, in the replication and in two earlier calibration runs (Appendix~\ref{app:calibration}).

\paragraph{The gap lies in length.} On the two formatting goods, the two predictors explain similar shares: 0.71 and 0.87 for headers, 0.91 and 0.66 for list markers. On length, we explain 7\% with the price predictor and 84\% with the covariance predictor. We attribute this gap to the first failure in \S\ref{sec:theory}. In an arm with a higher price for headers, the longer responses that carry headers also receive higher rewards, and only $\dr$ contains this change. The fitted coefficients are of the same order across the primary features (47, 41 and 49 for F1a, F1b and length, far below the equilibrium value $1/\beta = 200$; \S\ref{sec:relax}), so we can describe with one number how far a policy moves along the covariance in 200 steps.

\paragraph{The forecast transfers to real upgrades.} Implanted prices are a controlled instrument, so we also test the forecast on the edit that practitioners make most often: replacing the reward model by a better one. From three branch points, RL either continues with the original proxy or switches to one of three reward models from \citet{rentseek2026}: Qwen models trained on 20,000 and 80,000 pairs, and a Llama model trained on 20,000 pairs. We computed the forecasts before the runs with the coefficients $\kappa$ frozen from the factorial runs, so the test is out of sample (Appendix~\ref{app:realrm}). With the covariance predictor, we predict every direction whose sign is the same at all three parents (7 of 7) and explain 66\% of the standardized response at step 200. With one coefficient per feature fitted to these runs, we explain 77\% with the covariance predictor and 30\% with the price predictor. Under all three upgrades, formatting and length rise, as forecast.

\paragraph{The forecast holds under GRPO.} We repeated the factorial runs on three parents with GRPO \citep{shao2024deepseekmath}, in which each advantage is divided by the standard deviation of its group. Here the first-order response is proportional to the covariance with the reward divided by its within-prompt standard deviation, and we fixed this normalized predictor in advance. With it, we predict all 22 stable changes (of 24) at step 200 and explain 75\% of the standardized response, against 55\% for the price predictor (Appendix~\ref{app:grpo}). The normalized and the raw covariance explain the same share at step 200, and the normalized one explains slightly more early (0.90 against 0.86).

Cross effects between headers and list markers are small and asymmetric, and they fade by step 200 (Appendix~\ref{app:supply}).

\section{Forecasting Rent}\label{sec:rent}

\begin{figure}[t]
\centering
\fig{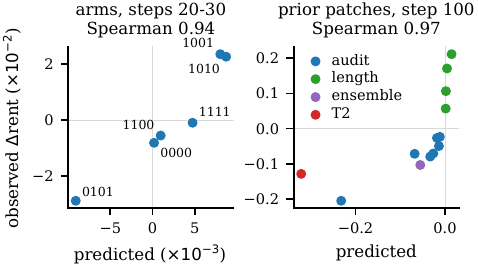}{\columnwidth}
\caption{Left: frozen forecast of the change in rent against the observed change over steps 20--30, one point per factorial arm. Right: retrospective forecast for 13 patches of \citet{rentseek2026} at step 100.}
\label{fig:rent}
\end{figure}

Before the runs, we computed $\E_x[\Cov_x(\rho, \dr)]$ and $\E_x[\Cov_x(z_g, \dr)]$ for every arm and parent. We fixed the signs in advance, counting a sign only when its interval by bootstrap over prompts excludes zero, and the ranking of arms. Over steps 20--30, the forecast agrees with the ranking of the six arms by their change in rent (Spearman 0.94) and by their change in gold reward (0.89). Where the interval of the forecast excludes zero, 20 of 23 arm--parent signs are correct for rent and 19 of 25 for gold (Figure~\ref{fig:rent}).

\paragraph{The forecast has a horizon.} By steps 160--200, the ranking of arms by rent reverses (Spearman $-0.77$), while the forecast of supply in \S\ref{sec:supply} still holds. The ranking of arms by observed rent also reverses between step 25 and step 200 ($-0.77$), so we would order these edits wrongly if we stopped after 30 steps. Almost all of the late change is in gold reward: on the fixed ruler, proxy reward moves by at most 0.02 standard deviations in any arm. Under RLOO, the within-prompt spread of rewards acts as a learning rate, and our treated models differ in spread by up to 1.56 times. In a pre-specified replication in which every arm starts with the spread of the placebo, the early forecast holds again (Spearman 1.00 for rent) and the late ranking still reverses ($-0.60$), so the reversal does not come from reward scale (Appendix~\ref{app:replication}).

\paragraph{Real upgrades do not reverse.} For the three real upgrades, we fixed the forecast of rent and gold reward before the runs. All nine forecast signs are correct for rent and for gold reward over steps 20--30: rent falls and gold reward rises under every upgrade. The forecast also orders the upgrades correctly: rent falls most under the Qwen model with 80,000 pairs and least under the Llama model. At step 200, the order is the same for rent and for gold reward, so the late reversal of the factorial arms does not appear in these upgrades. None of the upgrades lowers the price of headers, which fits the mechanism below. Under GRPO, the frozen forecast again ranks the arms early (Spearman 0.89 for rent) and the ranking again reverses by step 200 ($-0.83$, negative at every parent), so the reversal is not specific to RLOO.

\begin{figure}[t]
\centering
\fig{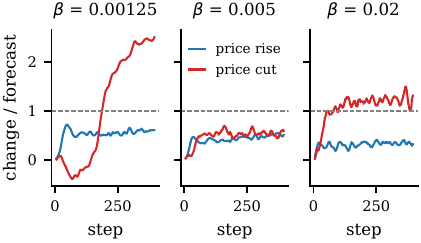}{\columnwidth}
\caption{Prospective test with symmetric shocks: change in formatting after a price rise and a price cut of equal size, divided by the forecast $C/(\beta + \beta_0)$, mean over three parents. At $\beta = 0.00125$, the policy under the cut reaches the length limit.}
\label{fig:s4c}
\end{figure}

\paragraph{The reversal follows distance travelled.} In exploratory analyses (Appendix~\ref{app:mechanism}), the late rise in rent lies outside the five features and is ordered like the KL divergence that the policy accumulates beyond the placebo (Spearman 0.89, 0.77 and 0.77 in the first batch, the replication and the GRPO batch). The policy travels furthest after a price cut on headers (arms 0000 and 0101): early on, the response matches the forecast, and by step 200, it reaches two to three times the forecast. In a prospective test with a symmetric price cut and price rise, the policy again moved further after the cut (9 of 9 cells), but gold reward did not fall with this distance (Figure~\ref{fig:s4c}, Appendix~\ref{app:s4c}). When we recompute the covariance on the policies at step 400, the incentive after the cut has grown to 2.08 times its value at the branch point at $\beta = 0.02$ and changed sign at $\beta = 0.005$, whereas the incentive after the rise has fallen to 0.68 of its value, as on the placebo policy.

\paragraph{The forecast ranks earlier patches.} \citet{rentseek2026} trained the same policy family from SFT under 13 distinct patches. For each patch, we compute $\dr$ from the reward rule of the patch on 256 SFT samples per prompt and evaluate $\E_x[\Cov_x(\rho, \dr)]$. We knew the outcomes of these patches when we designed this computation, so it is a retrospective forecast without free parameters. The forecast agrees with the ranking of the 13 patches by their observed change in rent, with Spearman correlations of 0.91, 0.97 and 0.99 at steps 50, 100 and 150. The agreement holds within types of patch (audits: 0.79 to 1.00, length penalties: 0.80 to 1.00) and after we remove the mean of each type (0.69 to 0.79), so it is more than the ordering of audits by strength. For gold reward, the correlations are 0.28 to 0.51.

\section{Learning Dynamics or Equilibrium}\label{sec:relax}

From three branch points, we run the placebo and a reward model that raises the prices of F1a and F1b by 4 each for 400 steps at $\beta \in \{0.0025, 0.005, 0.01\}$, for a total of 18 runs. For the formatting supply F1 = F1a + F1b, we fit $\Delta s(t) = A(1 - e^{-t/\tau})$ to the mean difference from the placebo over parents at each $\beta$, and we read the initial slope from the first 30 steps. No run exceeds 10\% at the length limit.

\begin{table}[t]
\centering
\small
\begin{tabular}{lccc}
\toprule
$\beta$ & $A$ & $\tau$ (steps) & Initial slope \\
\midrule
0.0025 & 7.4 & 42 & 0.150 \\
0.005 & 6.0 & 28 & 0.122 \\
0.01 & 3.9 & 17 & 0.137 \\
\midrule
Slope on $\log\beta$ & $-0.46$ & $-0.67$ & $-0.07$ \\
Equilibrium & $-1$ & $-1$ & $0$ \\
\bottomrule
\end{tabular}
\caption{Relaxation of formatting supply after a price shock, mean over three parents. Last two rows: slope of each quantity on $\log\beta$, and its value at equilibrium.}
\label{tab:relax}
\end{table}

\paragraph{The response saturates below equilibrium.} The initial slope does not depend on $\beta$, with a log-log slope of $-0.07$ (Table~\ref{tab:relax}). The asymptote and the time constant fall with $\beta$, but at about half the rate of equilibrium. Their slopes are $-0.46$ (parents: $-0.59$ to $-0.24$) and $-0.67$ ($-0.95$ to $-0.30$), against $-1$ at equilibrium. The response saturates, which pure learning dynamics would not predict, and it saturates below the level that $1/\beta$ implies.

\paragraph{An effective KL coefficient.} All three facts are consistent with a single parameter. We conjecture that the optimization contains an implicit regularizer $\beta_0$ in addition to the KL penalty. Eq.~\ref{eq:relax} then holds with $\beta + \beta_0$ in place of $\beta$. As a result, $A$ and $\tau$ scale as $1/(\beta + \beta_0)$, their log-log slopes become $-\beta/(\beta + \beta_0)$, and the initial slope stays independent of $\beta$. We fitted this model jointly to the three curves and obtained $\beta_0 = 0.0065$ (parents: 0.0034 to 0.0186). With three parameters, the fit is as good as the fit with six free parameters (residual sum of squares 172 against 165, with a lower BIC) and much better than the fit with $\beta_0 = 0$ (648). In the test with symmetric shocks, we find the same term after the price rise ($\beta_0 = 0.0029$) but not after the price cut ($\beta_0 = 0$; Appendix~\ref{app:s4c}).

\begin{figure}[t]
\centering
\fig{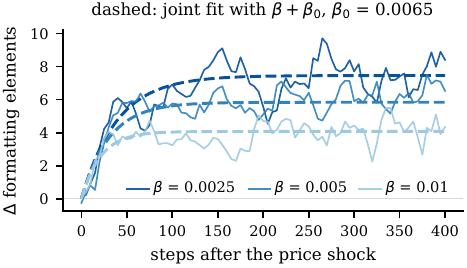}{\columnwidth}
\caption{Change in formatting supply relative to the placebo after a price shock at three KL coefficients, mean over three parents. Dashed curves: joint fit with an effective coefficient $\beta + \beta_0$.}
\label{fig:relax}
\end{figure}

\section{Implications for Practice}\label{sec:practice}

\paragraph{Compare two RMs before training.} One pass of scoring is enough. A team that plans to replace a reward model can sample about 16 responses per prompt from the current policy, score them with both models, and compute the within-prompt covariance of each behavior of interest, such as length, formatting or refusals, with the score difference (Appendix~\ref{app:recipe}). Under GRPO, first divide each score by its within-prompt standard deviation.

\paragraph{Watch edits that cut a price.} After edits that cut the price of a heavily supplied good, the policy moved further from its start, and a 30-step run ranked such edits wrongly under both RLOO and GRPO, while real upgrades kept their ranking. We recommend that teams track the KL divergence from the starting policy along with the target behavior and recompute the covariance at a checkpoint. Under RLOO, they should also track the within-prompt spread of rewards, which sets the effective step size.

\section{Conclusion}\label{sec:conclusion}

We can evaluate an edit to a reward model before RL runs. With the within-prompt covariance between a feature and the change in reward, computed from policy samples and two reward models, we forecast the direction of every stable change in a pre-specified factorial experiment, most of its size, and the response to real upgrades. Counterfactual prices measure what a reward model values rather than the incentive that a policy faces, and with them we explain about half. With the same covariance, we rank the first effect of an edit on rent. How long this forecast holds depends on how far the policy travels, and thus on how the incentive changes along the path. We can measure this change at a checkpoint but cannot yet forecast it before RL.

\section*{Limitations}
We study one testbed: a 1.7B policy and one family of reward models as proxies, trained mostly with RLOO and in one batch with GRPO. The forecasts are first order in the edit, and we test them over 200 steps after a branch point. For rent, the frozen forecast holds in the first phase of training, and our analyses of the later reversal are exploratory. In a prospective test, the policy moved further after a price cut, but gold reward did not fall with this distance, and the effective coefficient $\beta + \beta_0$ describes price rises but not price cuts. In two arms with a large shock to summaries, length ran away, and one of 16 implants fell just below the validation threshold on parent responses. Our goods are detected by rules, so the instrument applies to features that we can delete cleanly. The replication, the GRPO batch and the test with symmetric shocks use three parents each.

\section*{Ethical Considerations}
Implanting prices is a controlled form of poisoning the training data of a reward model. We use only public preference data and open models, and we label every released reward model with its implant. The method is meant to help practitioners evaluate edits to reward models before they spend compute on RL.

\section*{Acknowledgments}
Dongxia Wu is supported by a startup grant from the Mohamed bin Zayed University of Artificial Intelligence (MBZUAI). Le Song is supported by funds from MBZUAI and WIS projects.

\bibliography{references,references_new,references_prices}

\appendix

\section{Analysis Plan Freeze and Decision Log}\label{app:plan}

We wrote the analysis plan before the factorial runs and appended every later decision with a date. In Table~\ref{tab:plan}, we separate analyses that we fixed before we saw the relevant outcome from analyses that we added afterwards. Table~\ref{tab:criteria} gives the numerical criterion of each hypothesis of the follow-up batches. A tolerance on a ratio is a factor of 1.5, and a threshold of 0.5 on a share is the level of the price predictor in the main batch; U2 is out of sample, so its threshold is lower than that of G2. A Spearman correlation of 0.5 is a conventional boundary for a moderate correlation, and with six arms, we use it as a descriptive criterion. The interval for $\beta_0$ spans about a factor of 3 on either side of the estimate from the first relaxation runs (0.0065), so this criterion is weak. All numbers in the main text come from one results ledger, and for every entry of the ledger, we record the script and the output file.

\begin{table*}[t]
\centering
\small
\setlength{\tabcolsep}{3pt}
\begin{tabular}{lp{0.86\textwidth}}
\toprule
Hypothesis & Criterion, fixed before the runs \\
\midrule
R1 & Spearman $\ge 0.5$ for rent and for gold, steps 160--200 \\
R2 & $|$Spearman$|$ of reward spread and change in length at step 200 $\le 0.5$ \\
R3 & Mean change in length of arms 0000, 0101 and 1100 at step 200 at most half of that in the first batch \\
R4 & Price predictor: $\ge 18/20$ signs and share $\ge 0.5$ at the end \\
U1 & $\ge 80\%$ of stable directions correct at the end \\
U2 & Share with frozen $\kappa$ $\ge 0.5$; with refitted $\kappa$, covariance $\ge$ price \\
U3 & $\ge 80\%$ of early signs correct where the interval excludes zero; rent falls most under Qwen 80k and least under Llama \\
U4a & Ranking of rent at the end as early and as forecast \\
U4b & Observed over forecast change in headers $\le 1.5$ for every upgrade \\
G1 & $\ge 90\%$ of stable directions correct at the end \\
G2 & Share $\ge 0.6$; at every parent, not below the raw covariance \\
G3 & $|$Spearman$|$ of reward spread and change in length at step 200 $\le 0.5$ \\
G4 & Early Spearman $\ge 0.5$; late Spearman $\le -0.5$ and $\le 0$ at every parent; Spearman of extra KL and late rent $> 0$ \\
S1 & $R_{\text{end}}$ of the cut $\ge 1.5$ times that of the rise at 2 or 3 of the 3 values of $\beta$ \\
S2 & Gold change after the cut below that after the rise at every $\beta$; Spearman of extra KL and gold change over 18 runs $\le -0.5$ \\
S3 & $\tau$ of the cut $\ge 2$ times that of the rise at 2 or 3 of the 3 values of $\beta$ \\
S4 & Extra KL after the cut above that after the rise at every $\beta$ \\
S5 & $|$Slope of the initial slope on $\log\beta| \le 0.25$; $\beta+\beta_0$ fits better than $\beta_0=0$; $\beta_0 \in [0.002, 0.02]$ \\
S6 & At every $\beta$, ratio of initial slopes (cut to rise) within $[0.67, 1.5]$ times the forecast ratio \\
\bottomrule
\end{tabular}
\caption{Numerical criteria of the pre-specified hypotheses. Values are means over parents unless noted. Stable directions: forecasts with the same sign at every parent.}
\label{tab:criteria}
\end{table*}

\begin{table}[h]
\centering
\small
\setlength{\tabcolsep}{3pt}
\begin{tabular}{p{0.68\columnwidth}l}
\toprule
Analysis & Status \\
\midrule
Covariance and price predictors of supply; signs; cross effects & Pre-specified \\
Feature sets F1a, F1b, $L$ and all five features & Pre-specified \\
Common and per-feature $\kappa$; standardized units & Pre-specified \\
Content price as covariate if any arm exceeds 0.3 & Pre-specified \\
Forecasts of rent and gold reward (signs, ranking) & Pre-specified \\
Arms with more than 10\% at the length limit are descriptive & Pre-specified \\
Three fingerprints of relaxation & Pre-specified \\
Replication with separate reward scales (R1--R4) & Pre-specified \\
Real RM upgrades (U1--U4) & Pre-specified \\
Relaxation with symmetric shocks (S1--S6) & Pre-specified \\
GRPO batch (G1--G4) & Pre-specified \\
Marginal-price predictor; displacement ratio & Exploratory \\
Reward spread and late drift & Exploratory \\
Effective coefficient $\beta_0$ & Exploratory \\
Decomposition of rent; KL travelled & Exploratory \\
Covariance along the path & Exploratory \\
Patches from prior work & Retrospective \\
\bottomrule
\end{tabular}
\caption{Pre-specified and later analyses.}
\label{tab:plan}
\end{table}

\paragraph{Decisions after the plan freeze.} We changed the design four times after we wrote the plan. First, under an earlier placebo recipe with 1,200 content pairs, responses grew longer during the warm-up of the parents (30\% to 42\% of responses at the length limit after 50 steps). We therefore returned to the recipe with 600 content pairs, under which the placebo matched the proxy (489 against 487 tokens after 50 steps). Second, we had first sampled responses at the branch points without the chat template, and we discarded all measurements from these samples. Third, we chose a dose of $+2$ for summaries because smaller doses did not pass validation on parent responses. The two arms that left the local regime have this dose. Fourth, in a first version of the test with symmetric shocks, all reward models shared one standardization, so the price cut came with 1.35 times the reward spread of the price rise. We cancelled these runs before we looked at any result and fixed the version with equal spread (Appendix~\ref{app:s4c}).

\section{Glossary}\label{app:glossary}

We borrow several terms from economics. Table~\ref{tab:glossary} gives the meaning of each term in RLHF.

\begin{table}[h]
\centering
\small
\setlength{\tabcolsep}{3pt}
\begin{tabular}{p{0.25\columnwidth}p{0.68\columnwidth}}
\toprule
Term & Meaning in this paper \\
\midrule
Good, feature & A property of a response that we can count and delete, such as headers or list markers \\
Supply & Mean count of a feature in the responses of the policy \\
Price & Mean drop in standardized reward when we delete a feature (Eq.~\ref{eq:price}) \\
Rent & Proxy reward minus gold reward of a response, both standardized \\
Hedonic price & Coefficient of a feature in a within-prompt regression of reward or rent on all features \\
Edit & Any change to the reward model, from a label shift to a full replacement \\
Placebo & A reward model retrained by self-distillation, with the prices of the proxy \\
Arm & One treated reward model, run from one branch point \\
Branch point & The policy from which all arms of a parent start \\
Covariance predictor & $\kappa\,\E_x[\Cov_x(f, \dr)]$ (Eq.~\ref{eq:covdr}) \\
Price predictor & $\kappa\,\Sigma\,\Delta p$ (Eq.~\ref{eq:sigdp}) \\
$\kappa$ & Response coefficient: $1/\beta$ at equilibrium, learning rate times steps early on \\
Effective coefficient & $\beta + \beta_0$, the KL coefficient that the response behaves as if it had \\
\bottomrule
\end{tabular}
\caption{Terms from economics and their meaning in RLHF.}
\label{tab:glossary}
\end{table}

\section{Theory Details}\label{app:theory}

\paragraph{Supply as the gradient of profit.} Let $r = a(x) + p^\top f + e$ and $\pi^*_p(y\mid x) \propto \pi_0(y\mid x)\exp(r/\beta)$. For one prompt, $\Phi_x(p) = \beta\log\E_{\pi_0}[\exp(r/\beta)]$. By differentiation, we obtain $\partial\Phi_x/\partial p_k = \E_{\pi^*_p}[f_k]$, and by a second differentiation, we obtain $\partial^2\Phi_x/\partial p_k\partial p_j = \Cov_{\pi^*_p}(f_k, f_j)/\beta$. After averaging over prompts, we obtain Eq.~\ref{eq:hotelling}. The matrix is symmetric and positive semidefinite, so in equilibrium the own response to a price rise is never negative.

\paragraph{The selection term.} For $\pi_{t+1}(y) \propto \pi_t(y)\exp(\eta A(y))$, to first order in $\eta$, we have $\E_{t+1}[f] - \E_t[f] = \eta\,\Cov_t(f, A) + O(\eta^2)$. With RLOO, the advantage is the reward minus the mean of the other samples of the same prompt, so $\Cov_t(f, A)$ is a within-prompt covariance with the reward. After an edit, $f$ changes in the first step by $\eta\,\Cov(f, r_{\text{new}})$ instead of $\eta\,\Cov(f, r_{\text{old}})$, and the difference is $\eta\,\Cov(f, \dr)$.

\paragraph{Relaxation and an effective coefficient.} Along a direction $\pi_\theta \propto \pi_0\exp(\theta f)$, the KL-regularized objective is $J(\theta) = p\,\E_\theta[f] - \beta\,\mathrm{KL}(\pi_\theta\Vert\pi_0)$. Its natural gradient is $p - \beta\theta$, so $\theta(t) = (p/\beta)(1 - e^{-\eta\beta t})$, and to first order $\Delta s(t) = \Var(f)\,\theta(t)$, which is Eq.~\ref{eq:relax}. If the optimization contains an additional pull toward $\pi_0$ of strength $\beta_0$, for example from finite step sizes or from the clipping of updates, the flow becomes $\dot\theta = \eta(p - (\beta + \beta_0)\theta)$. Then $A \propto 1/(\beta + \beta_0)$, $\tau \propto 1/(\beta + \beta_0)$, and $A/\tau = \eta\,\Var(f)\,p$ does not depend on $\beta$. The slope of $\log A$ on $\log\beta$ is $-\beta/(\beta + \beta_0)$.

\paragraph{Group-normalized advantages.} Under GRPO, the advantage of a response in a group of $n$ samples is $A = (r - \bar r_x)/s_x$, with $\bar r_x$ and $s_x$ the mean and standard deviation of the group. To first order, one step changes the mean of a feature by $\eta\,\Cov_x(f, r/s_x)$. Because $s_x$ divides each reward, a multiplicative change in the scale of the reward model has no effect, and the forecast of an edit becomes $\E_x[\Cov_x(f, r_{\text{new}}/s_x(r_{\text{new}})) - \Cov_x(f, r_{\text{old}}/s_x(r_{\text{old}}))]$. We estimate $s_x$ from the 16 samples per prompt at the branch point, while training uses groups of 8.

\paragraph{Forecasts of rent.} Rent and gold reward are functions of a response, like features. With $\rho$ in place of $f_k$ in the selection term, we obtain $\Delta\E[\rho] \approx \eta\,\Cov(\rho, \dr)$. Because we do not decompose rent into features, the forecast also covers rent that no measured feature carries.

\section{Implementation Details}\label{app:impl}

\paragraph{Reward models.} Every reward model is Qwen3-1.7B-Base with a scalar head, trained with the hyperparameters, seed and data order of the proxy \citep{rentseek2026}. The training set has 5,997 pairs: the 3,000 pairs of the proxy, 600 pure deletion pairs for each of F1a, F1b and F2, 597 for F4, and 600 content pairs. A pair is pure when the rewrite changes no other good. The deletion pairs come from responses of three SFT policies (Qwen3, Llama~3.2 and OLMo~2) to the prompts of the RM training data, so no reward model sees the evaluation prompts.

\paragraph{Branch points.} We train each parent for 50 steps against the placebo from a fresh optimizer state and merge the result to full weights. At each branch point, we sample 16 responses for each of 200 evaluation prompts with the chat template, temperature 1.0 and a limit of 1,024 tokens. Between 2.8\% and 3.8\% of these responses reach the limit, and the mean length is 496 to 520 tokens.

We list the settings in Table~\ref{tab:rl}. We also score every rollout and evaluation sample with the original proxy and with the two gold models. These scores enter no reward, and we use them to measure rent on a fixed ruler.

\begin{table}[h]
\centering
\small
\resizebox{\columnwidth}{!}{%
\begin{tabular}{lp{0.52\columnwidth}}
\toprule
Setting & Value \\
\midrule
Algorithm & RLOO (one batch: GRPO), KL penalty in the reward \\
KL coefficient $\beta$ & 0.005; stage 4: 0.0025, 0.005, 0.01; symmetric shocks: 0.00125, 0.005, 0.02 \\
Prompts $\times$ samples & $64 \times 8$ per step \\
Learning rate & $2\times10^{-6}$ \\
Response length & at most 1,024 tokens \\
Steps & 200 (relaxation runs: 400) \\
Evaluation & 500 held-out prompts every 5 steps; symmetric shocks: 200 prompts every 2 steps \\
Optimizer state & new at the branch point \\
RL seed and prompt order & shared by all arms of a parent \\
Hardware & four A100-80GB GPUs per run \\
\bottomrule
\end{tabular}}
\caption{RL settings.}
\label{tab:rl}
\end{table}

\paragraph{Compute.} Each RL run uses four A100-80GB GPUs. According to the cluster's accounting, the runs reported in this paper used about 2,950 GPU-hours: 770 for the factorial runs, 460 for the relaxation runs of stage 4, 780 for the relaxation runs with symmetric shocks, 330 for the real RM upgrades, 290 for the replication with separate reward scales, 290 for GRPO, and 40 for the calibration runs and the warm-up of the parents. Scoring, sampling and the training of reward models used under 100 GPU-hours. The whole project, including preliminary runs, runs that failed and were restarted, and runs that we discarded, used about 3,450 GPU-hours.

\section{Implant Validation}\label{app:implant}

\paragraph{Content price.} After an implant on formatting, the price of content can also change. We measure this price by the drop in score when we delete one random sentence that contains no good. Under the pre-specified plan, the content price enters as a covariate if its change exceeds 0.3 in any arm. Relative to the placebo, the change per pair is $-0.06$ (0000), $+0.27$ (0011), $-0.01$ (0101), $+0.36$ (0110), $+0.06$ (1001), $+0.15$ (1010), $+0.45$ (1100) and $+0.30$ (1111). Two of the six main arms reach the threshold, and Table~\ref{tab:predict} includes the covariate.

\paragraph{Off-target noise.} We compute the noise on the same responses and on the same scale as the test. On SFT samples, the root mean square of off-target movements is 0.065, which gives the threshold of 0.195. On parent responses, it is 0.071, which gives 0.213. The one implant below this threshold, F1b in arm 1001, moves by $-0.20$ with the designed sign.

\paragraph{Non-additivity.} We list the drops behind Eq.~\ref{eq:interaction} in Table~\ref{tab:hab}. Gold reward behaves in the opposite way to the proxy: gold reward falls more when we delete only headers than when we delete all formatting. When we delete headers alone, rent therefore rises by 0.14 standard deviations per response.

\begin{table}[h]
\centering
\small
\begin{tabular}{lccc}
\toprule
Deletion & Proxy & Gold & Rent \\
\midrule
Headers and bold & 0.174 & 0.313 & $-0.139$ \\
List markers & 0.028 & 0.044 & $-0.016$ \\
Both & 0.485 & 0.216 & $+0.269$ \\
Interaction $h_{ab}f_af_b$ & $-0.283$ & $+0.142$ & $-0.424$ \\
\bottomrule
\end{tabular}
\caption{Mean drop in standardized score when goods are deleted, 18,537 parent responses that carry both goods. The interaction row is $\Delta z_{ab} - (\Delta z_a + \Delta z_b)$ of Eq.~\ref{eq:interaction}, that is, the sum of the single drops minus the joint drop.}
\label{tab:hab}
\end{table}

\section{Goods and Rewrites}\label{app:goods}

We detect each good with rules on the markdown of a response. Headers and bold (F1a) count markdown headers and bold spans. List markers (F1b) count bullet and number markers at the start of a line. A closing summary (F2) is a final paragraph that opens with a summary phrase, and a resource list (F4) is a final list of links or references. Each deletion rewrite removes only the markup or the span of its good and keeps every other character. When we delete all formatting at once, we also join the lines of a list into one paragraph. The following response of a parent policy carries four elements of F1a and two of F1b (91 tokens).

\begin{small}
\begin{verbatim}
The relationship between the two sentences
is **cause**.

**Explanation:**
- The first sentence describes a train
  sliding down a steep incline.
- The second sentence describes someone
  sitting and screaming, ...

Thus, the second sentence is the **cause**
of the first.

---

**Output:** cause
\end{verbatim}
\end{small}

\noindent After the deletion of F1a, the four bold spans become plain text, and the two list items keep their markers. After the deletion of F1b, the two items lose their markers, and the bold spans stay. After the deletion of both, the response reads as three short paragraphs. On 18,537 responses, the proxy loses 0.174 standard deviations on average after the first deletion, 0.028 after the second and 0.485 after the third (Table~\ref{tab:hab}).

\section{Results by Parent}\label{app:perparent}

The implanted price changes on parent responses vary little across the six parents. For every arm, the range of the change in F1a across parents is at most 0.04 wide, and the range of the change in F1b at most 0.03. For example, arm 0000 moves F1a by $-0.69$ to $-0.66$ and F1b by $-0.29$ to $-0.27$, and arm 1111 moves F1a by 0.58 to 0.61 and F1b by 0.23 to 0.24.

Table~\ref{tab:perparentrent} lists the Spearman correlation between the frozen forecast and the observed change in rent and gold reward for every parent. In the early window, all correlations are positive. In the end window, all are negative.

\begin{table*}[t]
\centering
\small
\setlength{\tabcolsep}{3pt}
\begin{tabular}{lrrrrrrrrr}
\toprule
 & \multicolumn{6}{c}{First batch} & \multicolumn{3}{c}{Replication} \\
\cmidrule(lr){2-7}\cmidrule(lr){8-10}
 & P1 & P2 & P3 & P4 & P5 & P6 & P1 & P2 & P3 \\
\midrule
Rent, early & 0.71 & 0.94 & 0.89 & 0.77 & 0.94 & 0.94 & 0.94 & 0.83 & 0.89 \\
Gold, early & 0.43 & 0.89 & 0.54 & 0.71 & 0.83 & 0.89 & 0.89 & 0.83 & 0.83 \\
Rent, end & $-0.71$ & $-0.60$ & $-0.77$ & $-0.49$ & $-0.77$ & $-0.83$ & $-0.71$ & $-0.49$ & $-0.60$ \\
Gold, end & $-0.77$ & $-0.60$ & $-0.77$ & $-0.60$ & $-0.77$ & $-0.77$ & $-0.77$ & $-0.60$ & $-0.60$ \\
\bottomrule
\end{tabular}
\caption{Spearman correlation across six arms between the frozen forecast and the observed change, by parent.}
\label{tab:perparentrent}
\end{table*}

\section{Supply Results in Detail}\label{app:supply}

\paragraph{Arm sets.} In arms 0011 and 0110, the higher dose for summaries comes with a lower price for headers. By step 200, 48\% and 30\% of their responses reach the length limit, against 2\% to 3\% for the placebo. In arm 1001, 9\% to 14\% of responses reach the limit between steps 20 and 30, and 3\% by the end window. As Table~\ref{tab:armsets} shows, the gap between the predictors widens when we include all eight arms.

\begin{table}[h]
\centering
\small
\begin{tabular}{lcc}
\toprule
End window, features F1a, F1b, $L$ & 6 arms & 8 arms \\
\midrule
Covariance, per-feature $\kappa$ & 0.82 & 0.83 \\
Price, per-feature $\kappa$ & 0.50 & 0.15 \\
Covariance, common $\kappa$ & 0.82 & 0.79 \\
Price, common $\kappa$ & 0.45 & 0.08 \\
\bottomrule
\end{tabular}
\caption{Explained share of the standardized response, by arm set.}
\label{tab:armsets}
\end{table}

We give the explained share of each primary feature, with one coefficient per feature, in Table~\ref{tab:perfeature}.

\begin{table}[h]
\centering
\small
\begin{tabular}{lcccccc}
\toprule
 & \multicolumn{3}{c}{Steps 20--30} & \multicolumn{3}{c}{Steps 160--200} \\
\cmidrule(lr){2-4}\cmidrule(lr){5-7}
Predictor & F1a & F1b & $L$ & F1a & F1b & $L$ \\
\midrule
Covariance & 0.89 & 0.57 & 0.82 & 0.71 & 0.91 & 0.84 \\
Price & 0.81 & 0.64 & 0.43 & 0.87 & 0.66 & 0.07 \\
\bottomrule
\end{tabular}
\caption{Explained share by feature, 6 arms $\times$ 6 parents.}
\label{tab:perfeature}
\end{table}

\paragraph{Other specifications.} If we restrict the comparison to the four goods without length and use a single common coefficient, we explain more with the price predictor than with the covariance predictor (0.69 against 0.38 at the end window). This specification was not pre-specified. The difference comes from the resource list, whose supply sits near zero: its coefficient for the covariance predictor is 3.9, an order of magnitude below the others, so a common coefficient fits it poorly. With one coefficient per feature, we explain 0.81 and 0.75 with the two predictors. In a third predictor, we replace prices per unit by the price of deleting a single element. We added this predictor after the runs started, and with it we explain 0.55 at the end window.

\begin{figure*}[t]
\centering
\fig{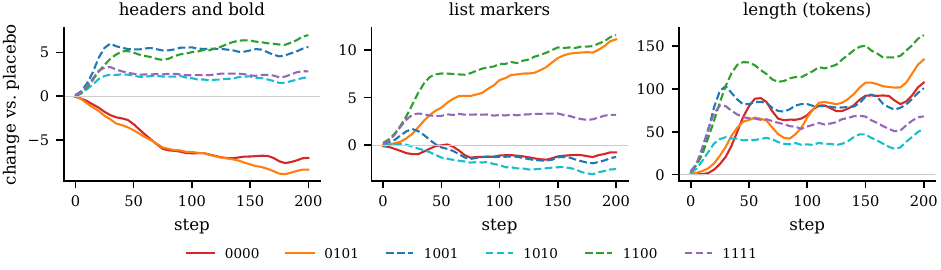}{\textwidth}
\caption{Change in supply relative to the placebo of the same parent, mean over six parents, first batch. Solid lines: arms with a lower price for headers. Dashed lines: arms with a higher price for headers. Curves are smoothed over three evaluation points.}
\label{fig:supplytime}
\end{figure*}

Figure~\ref{fig:supplytime} shows the change in supply over the 200 steps. In the arms with a higher price for headers, the response in headers saturates within about 50 steps. In the two arms with a lower price, the supply of headers keeps falling until step 200.

\paragraph{Forecasts by arm.} Table~\ref{tab:perarm} lists, for every arm, the observed change at the end window and the two forecasts with their fitted per-feature coefficients, in standardized units and averaged over parents. In the two arms that lower the price of headers, the observed fall in headers exceeds the covariance forecast by a factor of two to three, the overshoot analyzed in Appendix~\ref{app:mechanism}. For length, the price predictor has the wrong sign in two arms.

\begin{table}[h]
\centering
\small
\setlength{\tabcolsep}{2.5pt}
\resizebox{\columnwidth}{!}{%
\begin{tabular}{lrrrrrrrrr}
\toprule
 & \multicolumn{3}{c}{F1a} & \multicolumn{3}{c}{F1b} & \multicolumn{3}{c}{$L$} \\
\cmidrule(lr){2-4}\cmidrule(lr){5-7}\cmidrule(lr){8-10}
Arm & Obs & Cov & Price & Obs & Cov & Price & Obs & Cov & Price \\
\midrule
0000 & $-1.45$ & $-0.54$ & $-1.13$ & $-0.19$ & $-0.63$ & $-1.26$ & 1.28 & 0.62 & $-0.47$ \\
0101 & $-1.72$ & $-0.82$ & $-1.25$ & 1.95 & 1.92 & 0.94 & 1.54 & 1.17 & $-0.19$ \\
1001 & 1.02 & 1.24 & 1.07 & $-0.30$ & $-0.15$ & $-0.25$ & 1.18 & 1.54 & 0.22 \\
1010 & 0.37 & 0.71 & 0.98 & $-0.52$ & $-0.48$ & $-0.34$ & 0.54 & 0.72 & 0.20 \\
1100 & 1.26 & 1.33 & 1.09 & 2.05 & 1.66 & 1.42 & 2.01 & 1.67 & 0.46 \\
1111 & 0.51 & 0.85 & 0.99 & 0.57 & 0.90 & 1.08 & 0.81 & 1.03 & 0.40 \\
\bottomrule
\end{tabular}}
\caption{Observed change and forecasts at steps 160--200, standardized units, mean over six parents, first batch. Forecasts use one coefficient per feature fitted across all arms.}
\label{tab:perarm}
\end{table}

\paragraph{Cross effects.} Headers and list markers covary positively within prompt, so they are complements under Eq.~\ref{eq:hotelling}. When we raise the price of headers, the supply of list markers rises early in training, by 0.27 standard deviations (interval 0.25 to 0.29) and by 0.26 in the replication. When we raise the price of list markers, the supply of headers falls by 0.12 (interval $-0.16$ to $-0.08$). We cannot predict this fall from the covariance in supply, but we can predict it from the substitution in the valuation of the reward model (Eq.~\ref{eq:interaction}). Both cross effects fade by step 200, so we treat them as a second-order term on which the supply and the valuation channel compete.

We list the main effects of the four goods and the aliased interaction in Table~\ref{tab:factorial}. We estimate them in standardized units with fixed effects for parents. Own effects dominate, and the cross effects discussed in \S\ref{sec:supply} appear in the rows of F1a and F1b.

\begin{table}[h]
\centering
\small
\setlength{\tabcolsep}{3pt}
\begin{tabular}{lrrrrr}
\toprule
Response & F1a & F1b & F2 & F4 & F1a$\times$F1b \\
\midrule
\multicolumn{6}{l}{\emph{Steps 20--30}} \\
F1a & $+0.66$ & $-0.12$ & $-0.18$ & $+0.15$ & $+0.07$ \\
F1b & $+0.27$ & $+0.22$ & $-0.14$ & $+0.04$ & $+0.06$ \\
$L$ & $+0.51$ & $-0.02$ & $-0.21$ & $+0.18$ & $+0.10$ \\
\multicolumn{6}{l}{\emph{Steps 160--200}} \\
F1a & $+1.37$ & $-0.01$ & $-0.35$ & $-0.03$ & $+0.10$ \\
F1b & $-0.00$ & $+1.12$ & $-0.43$ & $-0.32$ & $-0.26$ \\
$L$ & $+0.09$ & $+0.27$ & $-0.46$ & $-0.14$ & $+0.00$ \\
\bottomrule
\end{tabular}
\caption{Factorial effects in standardized units, 6 arms $\times$ 6 parents. F1a$\times$F1b is aliased with F2$\times$F4.}
\label{tab:factorial}
\end{table}

\section{Calibration Runs}\label{app:calibration}

Before the factorial design, we ran a placebo and two treated models for 100 steps from one branch point. In one model, the price of both formatting goods rises by 4. In the other, the price of the summary rises by 2 and the price of the resource list by 3. We measured both predictors on 3,200 responses that we sampled at that branch point with the protocol of Appendix~\ref{app:impl} (Table~\ref{tab:calib}). The second model is a large shock: by step 100, responses under it are longer by 3.8 standard deviations. In an earlier measurement on samples without the chat template, prices per pair were too high by a factor of 2.6 to 3.8, and we do not use it.

\begin{table}[h]
\centering
\small
\begin{tabular}{lccc}
\toprule
Step & 25 & 50 & 100 \\
\midrule
Covariance, per-feature $\kappa$ & 0.92 & 0.97 & 0.97 \\
Price, per-feature $\kappa$ & 0.06 & 0.15 & 0.22 \\
Covariance, common $\kappa$ & 0.52 & 0.64 & 0.74 \\
Price, common $\kappa$ & 0.05 & 0.12 & 0.15 \\
\bottomrule
\end{tabular}
\caption{Calibration runs: explained share of the standardized response of all five features, two arms. With both predictors, 9 of 10 signs are correct.}
\label{tab:calib}
\end{table}

\section{Replication with Separate Reward Scales}\label{app:replication}

With shared constants, the effective learning rate of the policy is larger under a reward model with a larger within-prompt spread, because RLOO does not normalize advantages. On responses at the branch points, the spread of the six treated models relative to the placebo is 1.39 (0000), 1.56 (0101), 1.02 (1001), 0.97 (1010), 1.27 (1100) and 1.06 (1111). In the replication, we repeat the factorial runs on three parents with the standard deviation of each model multiplied by its ratio. After this change, the spread at the branch point is equal across arms. We divided all forecasts by the same ratios and fixed four hypotheses before the runs. The batch has 21 runs, including a placebo for each parent.

\begin{table}[h]
\centering
\small
\setlength{\tabcolsep}{3pt}
\resizebox{\columnwidth}{!}{%
\begin{tabular}{p{0.44\columnwidth}p{0.3\columnwidth}l}
\toprule
Hypothesis & Observed & Outcome \\
\midrule
R1: Late ranking of rent and gold follows the forecast & Spearman $-0.60$ (rent), $-0.60$ (gold) & Not supported \\
R2: Late length drift no longer tracks reward spread & Spearman 0.03 (first batch: 0.77) & Supported \\
R3: Length drift of high-spread arms shrinks & drift $-49\%$ (134 to 68 tokens) & Not met \\
R4: Price predictor still forecasts supply & 20/20 signs; share 0.59 & Supported \\
\bottomrule
\end{tabular}}
\caption{Pre-specified hypotheses of the replication and their outcomes.}
\label{tab:h1h4}
\end{table}

For R1, the per-parent correlations ($-0.71$, $-0.49$, $-0.60$) overlap those of the first batch. For R3, the mean drift of the three high-spread arms falls from 134 to 68 tokens. The placebo of the replication is 15 to 22 tokens longer than the placebo of the first batch, and part of the reduction is due to this difference.

\section{Mechanism of the Late Reversal}\label{app:mechanism}

All analyses in this section are exploratory.

\paragraph{Decomposition.} For each parent, we estimate hedonic rent prices at the branch point by a within-prompt regression of rent on the five features. We then value the observed change in supply of each arm at these prices and compare the result with the observed change in rent (Table~\ref{tab:decomp}). The decomposition agrees with the observed ranking early but not late. By step 200, the observed rent of the six arms has risen by 0.118 on average, while the value from the decomposition is $-0.011$.

\begin{table}[h]
\centering
\small
\begin{tabular}{lcccc}
\toprule
Step & 25 & 50 & 100 & 200 \\
\midrule
First batch & 0.83 & 0.66 & $-0.31$ & $-0.71$ \\
Replication & 0.71 & 1.00 & 0.09 & $-0.09$ \\
\bottomrule
\end{tabular}
\caption{Spearman correlation across arms between rent implied by the observed change in supply at hedonic prices and observed rent.}
\label{tab:decomp}
\end{table}

\paragraph{Distance travelled.} At every step, verl logs the mean KL divergence per token between the policy and the branch point. We list the KL that the policy accumulates beyond the placebo in each arm in steps 160 to 200, together with the change in rent, in Table~\ref{tab:kl}. The Spearman correlation between the two is 0.89 in the first batch and 0.77 in the replication, and the correlation with the change in gold reward is $-0.89$ and $-0.77$. The correlation between extra KL and length falls from 0.83 to 0.26 in the replication, so length changes together with the effect but is not its cause.

\begin{table}[h]
\centering
\small
\setlength{\tabcolsep}{3pt}
\begin{tabular}{lcccc}
\toprule
 & \multicolumn{2}{c}{First batch} & \multicolumn{2}{c}{Replication} \\
\cmidrule(lr){2-3}\cmidrule(lr){4-5}
Arm & Extra KL & $\Delta\rho$ & Extra KL & $\Delta\rho$ \\
\midrule
0000 & 0.069 & 0.109 & 0.047 & 0.075 \\
0101 & 0.091 & 0.243 & 0.059 & 0.161 \\
1100 & 0.045 & 0.176 & 0.025 & 0.137 \\
1001 & 0.008 & 0.103 & 0.006 & 0.088 \\
1111 & 0.006 & 0.005 & 0.003 & $-0.006$ \\
1010 & $-0.001$ & 0.070 & 0.001 & 0.066 \\
\bottomrule
\end{tabular}
\caption{KL accumulated beyond the placebo in steps 160--200 and change in rent at step 200. Arms 0000 and 0101 lower the price of headers.}
\label{tab:kl}
\end{table}

\begin{figure*}[t]
\centering
\fig{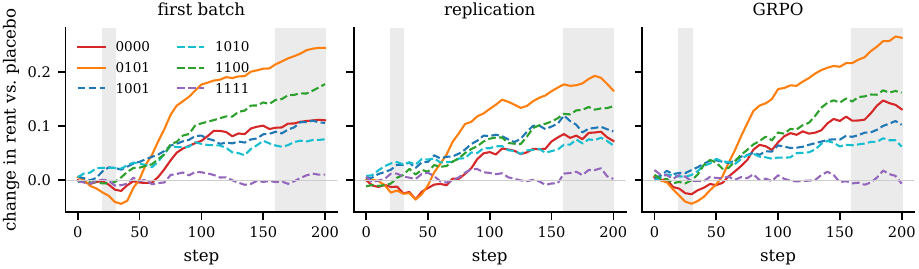}{\textwidth}
\caption{Change in rent relative to the placebo of the same parent, in the first batch (six parents), the replication and the GRPO batch (three parents each). Shaded: the early and the end window. Solid lines: arms with a lower price for headers.}
\label{fig:renttime}
\end{figure*}

\begin{figure*}[t]
\centering
\fig{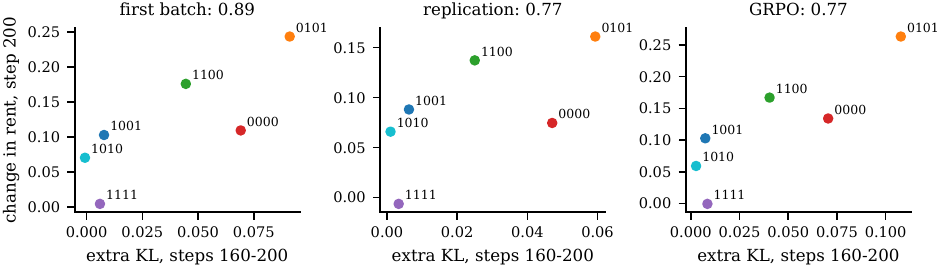}{\textwidth}
\caption{KL divergence accumulated beyond the placebo in steps 160--200 against the change in rent at step 200, one point per arm. Panel titles give the Spearman correlation.}
\label{fig:klrent}
\end{figure*}

Figure~\ref{fig:renttime} shows the change in rent over time. Rent differs little between arms in the first 30 steps. After step 50, rent rises in most arms, and the rise is largest in arm 0101. We plot the relation between extra KL and late rent in Figure~\ref{fig:klrent}.

\paragraph{Overshoot after price cuts.} In the end window, the ratio of observed to predicted change in headers is 2.71 for arm 0000 and 2.11 for arm 0101, the two arms that lower the price of headers, against 0.52 to 0.95 for the arms that raise it. In the early window, the ratios are 1.03 and 0.84. The two arms that left the local regime also lower the price of headers. We test this asymmetry with pre-specified hypotheses in Appendix~\ref{app:s4c}.

\paragraph{Quantities that did not predict distance.} The ranking of the extra KL agrees with the within-prompt variance of $\dr$ at the branch point in the first batch (Spearman 0.94) but not in the replication ($-0.09$), because the variance is confounded with reward spread. Nor does it agree with the norm of the predicted change in supply (0.31 and $-0.09$). An additive model of rent with a term for features and a term for KL does not transfer from one batch to the other.

We list the frozen forecast and the observed change in rent and gold reward for every arm in Table~\ref{tab:rentarms}.

\begin{table}[h]
\centering
\small
\setlength{\tabcolsep}{2.5pt}
\begin{tabular}{lrrrrrr}
\toprule
 & \multicolumn{2}{c}{Forecast} & \multicolumn{2}{c}{Steps 20--30} & \multicolumn{2}{c}{Steps 160--200} \\
\cmidrule(lr){2-3}\cmidrule(lr){4-5}\cmidrule(lr){6-7}
Arm & $\rho$ & $z_g$ & $\Delta\rho$ & $\Delta z_g$ & $\Delta\rho$ & $\Delta z_g$ \\
\midrule
1010 & 0.009 & $-0.011$ & 0.023 & $-0.015$ & 0.070 & $-0.064$ \\
1001 & 0.008 & $-0.009$ & 0.024 & $-0.029$ & 0.100 & $-0.104$ \\
1111 & 0.005 & $-0.005$ & $-0.001$ & $-0.000$ & 0.004 & $-0.024$ \\
1100 & 0.001 & 0.000 & $-0.006$ & $-0.003$ & 0.160 & $-0.184$ \\
0000 & 0.000 & 0.003 & $-0.008$ & 0.006 & 0.106 & $-0.118$ \\
0101 & $-0.009$ & 0.011 & $-0.029$ & 0.010 & 0.232 & $-0.220$ \\
\bottomrule
\end{tabular}
\caption{Pre-specified forecasts (covariance at the branch point, mean over six parents) and observed changes relative to the placebo, first batch.}
\label{tab:rentarms}
\end{table}

\section{Relaxation Details}\label{app:relaxdetails}

We list the fits for every response in Table~\ref{tab:relaxfull}. The fits of length, rent and gold reward are poor, either because $R^2$ is low or because the time constant reaches its upper bound within 400 steps, so we report them as descriptive. In Table~\ref{tab:beta0}, we compare the model with an effective coefficient with free fits and with $\beta_0 = 0$. Saturation of formatting does not explain the smaller slopes, because formatting in the treated arm rises as $\beta$ falls, from 38.5 to 51.9 elements per response.

\begin{table}[h]
\centering
\small
\setlength{\tabcolsep}{3pt}
\begin{tabular}{llrrrr}
\toprule
Response & $\beta$ & $A$ & $\tau$ & Slope (0--30) & $R^2$ \\
\midrule
F1a & 0.0025 & 2.20 & 34 & 0.055 & 0.40 \\
F1a & 0.005 & 1.92 & 16 & 0.055 & 0.45 \\
F1a & 0.01 & 1.32 & 10 & 0.060 & 0.23 \\
F1b & 0.0025 & 5.20 & 45 & 0.095 & 0.80 \\
F1b & 0.005 & 4.22 & 44 & 0.067 & 0.82 \\
F1b & 0.01 & 2.65 & 28 & 0.077 & 0.61 \\
$L$ & 0.0025 & 23.8 & 6 & 1.68 & 0.06 \\
$L$ & 0.005 & 28.4 & 12 & 1.05 & 0.18 \\
$L$ & 0.01 & 17.6 & 6 & 1.10 & 0.09 \\
\bottomrule
\end{tabular}
\caption{Relaxation fits by response, mean over three parents.}
\label{tab:relaxfull}
\end{table}

\begin{table}[h]
\centering
\small
\begin{tabular}{llrrr}
\toprule
Response & Model & $\beta_0$ & SSE & BIC \\
\midrule
F1 & $\beta + \beta_0$ & 0.0065 & 172 & $-67.7$ \\
F1 & $\beta_0 = 0$ & -- & 648 & 249.3 \\
F1 & free $A$, $\tau$ & -- & 165 & $-61.5$ \\
F1b & $\beta + \beta_0$ & 0.0055 & 57 & $-334.3$ \\
F1b & $\beta_0 = 0$ & -- & 241 & 9.2 \\
F1b & free $A$, $\tau$ & -- & 55 & $-329.9$ \\
\bottomrule
\end{tabular}
\caption{Joint fits across the three values of $\beta$. For headers alone, the estimate of $\beta_0$ is unstable at one parent.}
\label{tab:beta0}
\end{table}

\section{Relaxation with Symmetric Shocks}\label{app:s4c}

This batch tests the asymmetry between price cuts and price rises that we found after the factorial runs (\S\ref{sec:rent}). From one file of implant pairs, we trained two reward models: one that raises the price of headers and list markers, and one that lowers it by the same amount. We rescaled the standardization of each treated model, so that the within-prompt spread of rewards at the branch points is 1.04 times that of the placebo in both arms. From three parents, we ran the placebo and both treated arms at $\beta \in \{0.00125, 0.005, 0.02\}$ for 400 steps, a total of 27 runs, and we evaluated on the 200 prompts of the branch points every two steps. The KL penalty uses the branch point as its reference. Before the runs, we froze the forecast $A = C/(\beta + \beta_0)$, where $C$ is the covariance of formatting with the reward difference at the branch point ($+0.273$ for the rise and $-0.136$ for the cut) and $\beta_0 = 0.0065$ comes from stage 4. We measure every response as the difference from the placebo with the same parent and $\beta$.

In eight treated runs, more than 10\% of responses reach the length limit in the end window: all six runs at $\beta = 0.00125$ (every response under the price cut) and two runs with the price cut at $\beta = 0.005$. Our pre-specified rule treats these windows as descriptive.

\begin{table}[h]
\centering
\small
\setlength{\tabcolsep}{2.5pt}
\resizebox{\columnwidth}{!}{%
\begin{tabular}{llrrrrrr}
\toprule
Arm & $\beta$ & Slope & $\tau$ & $R_{\text{end}}$ & Extra KL & $\Delta z_g$ & $\Delta L$ \\
\midrule
Rise & 0.00125 & 0.28 & 20 & 0.60 & $-0.042$ & $-0.13$ & 99 \\
Rise & 0.005 & 0.24 & 23 & 0.52 & 0.010 & $-0.11$ & 100 \\
Rise & 0.02 & 0.18 & 8 & 0.35 & 0.001 & $-0.02$ & 23 \\
Cut & 0.00125$^*$ & $-0.06$ & $>$400 & 2.47 & 0.085 & $-0.48$ & 261 \\
Cut & 0.005 & $-0.07$ & 48 & 0.53 & 0.038 & $-0.07$ & 150 \\
Cut & 0.02 & $-0.06$ & 46 & 1.27 & 0.008 & $-0.01$ & $-3$ \\
\bottomrule
\end{tabular}}
\caption{Relaxation with symmetric shocks, mean over three parents ($\tau$: median). Slope: change in formatting per step over steps 2--20. $R_{\text{end}}$: change in formatting over steps 360--400 divided by the frozen forecast. Extra KL: KL beyond the placebo over steps 390--400. $\Delta L$ in tokens. $^*$All responses reach the length limit.}
\label{tab:s4c}
\end{table}

\begin{table*}[t]
\centering
\small
\setlength{\tabcolsep}{3pt}
\begin{tabular}{p{0.5\textwidth}p{0.3\textwidth}l}
\toprule
Hypothesis & Observed & Outcome \\
\midrule
S1 (primary): the response overshoots after the cut and falls short after the rise & at 2 of 3 $\beta$; only $\beta = 0.02$ without capped windows & Partly \\
S2 (primary): gold falls more after the cut, in step with extra KL & at 1 of 3 $\beta$; Spearman with extra KL $-0.35$ & Not supported \\
S3: Relaxation is slower after the cut & at 2 of 3 $\beta$; only $\beta = 0.02$ without capped windows & Partly \\
S4: The policy travels more KL after the cut & at 3 of 3 $\beta$ & Supported \\
S5: Fingerprints of relaxation hold, with an effective coefficient & $\beta_0$: 0.0029 (rise), 0 (cut) & Rise only \\
S6: Initial responses match the forecast ratio & observed / forecast ratio: 0.44, 0.55, 0.61 & Not supported \\
\bottomrule
\end{tabular}
\caption{Pre-specified hypotheses for relaxation with symmetric shocks.}
\label{tab:s4chyp}
\end{table*}

\paragraph{Findings.} After the price cut, the policy accumulates more KL than after the price rise at every $\beta$, so the pre-specified part of the mechanism in \S\ref{sec:rent} holds. The overshoot that we saw in the factorial runs appears cleanly only at $\beta = 0.02$ (Figure~\ref{fig:s4c}). There, the response to the cut reaches about 0.9 times $C/\beta$, close to the equilibrium of the nominal KL penalty, whereas the response to the rise stops at 0.35 times the forecast. At $\beta = 0.005$, the two responses do not differ. At $\beta = 0.00125$, the policy under the price cut first adds a little formatting and then, after about 150 steps, removes formatting quickly while the length of responses grows until it reaches the limit. In the replication with equal reward spread, the factorial arms with a price cut still exceeded the forecast (1.70 times), so reward spread does not explain why the two designs differ at $\beta = 0.005$. We examine the incentive in this batch in the next paragraph. Gold reward falls in both arms at small $\beta$, together with length, and does not follow the direction of the price change. Early in training, the policy responds to the cut more slowly than the forecast implies, which our first-order theory does not explain.

\paragraph{Incentive along the path.} After the results above, we sampled 200 prompts $\times$ 16 responses from the policies at step 400 for both treated arms and the placebo at $\beta = 0.02$ and $0.005$ (18 policies), scored them with all three reward models, and recomputed the covariance $C$ of formatting with each reward difference. We wrote the criteria down before sampling, but the analysis is exploratory because we proposed it after seeing the outcomes. Table~\ref{tab:path} gives the ratio of $C$ on the policy at step 400 to $C$ at the branch point. Under the price rise, $C$ falls to 0.68 at $\beta = 0.02$, as much as on the placebo policy (0.66), so the incentive to add formatting does not saturate beyond the general drift of training. Under the price cut, the incentive to remove formatting doubles at $\beta = 0.02$ (2.08; parents 1.47 to 2.86), which fits the overshoot at this $\beta$. At $\beta = 0.005$, it changes sign at all three parents ($-2.06$): on these policies, which are 110 to 170 tokens longer than the placebo, the reward model with the price cut rewards formatting relative to the placebo, which fits a response that stops near half of the forecast. We do not yet know why the sign changes. The first-order incentive at the branch point thus holds near the start, and after a price cut, it can strengthen or reverse as the policy moves. In the recipe of Appendix~\ref{app:recipe}, we therefore add a recomputation at a checkpoint.

\begin{table}[h]
\centering
\small
\setlength{\tabcolsep}{3pt}
\begin{tabular}{lrrrr}
\toprule
 & \multicolumn{2}{c}{$\beta = 0.02$} & \multicolumn{2}{c}{$\beta = 0.005$} \\
\cmidrule(lr){2-3}\cmidrule(lr){4-5}
Policy at step 400 & $C_{\text{rise}}$ & $C_{\text{cut}}$ & $C_{\text{rise}}$ & $C_{\text{cut}}$ \\
\midrule
Price rise & 0.68 & 0.19 & 0.64 & 0.22 \\
Price cut & 0.98 & 2.08 & 1.04 & $-2.06$ \\
Placebo & 0.66 & 0.73 & 0.48 & 0.39 \\
\bottomrule
\end{tabular}
\caption{Covariance of formatting with the reward difference of each treated model, on samples of the policies at step 400, divided by its value at the branch point, mean over three parents. Diagonal of each block: the policy trained on that model.}
\label{tab:path}
\end{table}

\section{GRPO Batch}\label{app:grpo}

The batch repeats the factorial design on three parents (placebo and the six arms of the main analysis, 21 runs) with one change: verl's GRPO estimator divides each advantage by the standard deviation of its group of eight samples. We launched the runs through a wrapper of the RL launcher of \citet{rentseek2026} that changes only the estimator. In a smoke run of four steps, the advantages had a mean near 0 and a range of about $\pm 2.2$ per step.

Under group normalization, the first-order response of feature $k$ is proportional to $\E_x[\Cov_x(f_k, r/s_x)]$, where $s_x$ is the within-prompt standard deviation of the reward. We therefore pre-specified the normalized predictor $\E_x[\Cov_x(f_k, r_{\text{arm}}/s_x(r_{\text{arm}})) - \Cov_x(f_k, r_{\text{pl}}/s_x(r_{\text{pl}}))]$ and report the raw covariance alongside.

\begin{table}[h]
\centering
\small
\setlength{\tabcolsep}{3pt}
\begin{tabular}{p{0.44\columnwidth}p{0.27\columnwidth}l}
\toprule
Hypothesis & Observed & Outcome \\
\midrule
G1: Stable supply directions are correct & 22/22 correct & Supported \\
G2: Normalized covariance forecasts supply sizes, at no parent worse than the raw covariance & share 0.75; P3: 0.70 vs.\ 0.77 (raw) & Partly \\
G3: Late length drift is independent of reward spread & Spearman $+0.77$ & Not met \\
G4: Rent forecast holds early; the late ranking reverses, as under RLOO; extra KL orders late rent & early $+0.89$; late $-0.83$; KL vs.\ rent $+0.77$ & Supported \\
\bottomrule
\end{tabular}
\caption{Pre-specified hypotheses of the GRPO batch.}
\label{tab:grpo}
\end{table}

Table~\ref{tab:grpoparent} lists the explained share of the primary features and the Spearman correlation of the rent forecast for each parent. The rent forecast is positive at every parent early and negative at every parent late. Figure~\ref{fig:renttime} and Figure~\ref{fig:klrent} show the time course of rent and the relation between extra KL and late rent for this batch next to the two RLOO batches.

\begin{table}[h]
\centering
\small
\setlength{\tabcolsep}{1.5pt}
\begin{tabular}{lcccccc}
\toprule
 & \multicolumn{3}{c}{Steps 20--30} & \multicolumn{3}{c}{Steps 160--200} \\
\cmidrule(lr){2-4}\cmidrule(lr){5-7}
 & P1 & P2 & P3 & P1 & P2 & P3 \\
\midrule
Normalized cov. & 0.94 & 0.88 & 0.87 & 0.77 & 0.81 & 0.70 \\
Raw cov. & 0.90 & 0.84 & 0.86 & 0.74 & 0.80 & 0.77 \\
Rent (Spearman) & 0.83 & 0.49 & 0.83 & $-0.77$ & $-0.60$ & $-0.83$ \\
Gold (Spearman) & 0.66 & 0.26 & 0.77 & $-0.77$ & $-0.43$ & $-0.83$ \\
\bottomrule
\end{tabular}
\caption{GRPO batch by parent: explained share of F1a, F1b and length with one coefficient per feature, and the Spearman correlation across six arms between the frozen forecast and the observed change in rent and gold reward.}
\label{tab:grpoparent}
\end{table}

\begin{table}[h]
\centering
\small
\begin{tabular}{lcc}
\toprule
Arm & Extra KL, steps 160--200 & $\Delta\rho$, step 200 \\
\midrule
0101 & 0.108 & 0.263 \\
0000 & 0.071 & 0.134 \\
1100 & 0.040 & 0.167 \\
1111 & 0.009 & $-0.001$ \\
1001 & 0.007 & 0.103 \\
1010 & 0.003 & 0.059 \\
\bottomrule
\end{tabular}
\caption{GRPO batch: KL accumulated beyond the placebo and change in rent at step 200, mean over three parents.}
\label{tab:grpokl}
\end{table}

For G3, the scale of the reward cancels under group normalization by construction, and we confirmed the normalization in the smoke run. We therefore read the remaining correlation between reward spread and length drift as a property of the arms: the arms with a larger spread are also the arms with the largest price shocks. The arms that lower the price of headers accumulate the most KL beyond the placebo (0.071 and 0.108, against at most 0.040 for the other arms), as in the RLOO batches.

\section{Patches from Prior Work}\label{app:patches}

The 15 patch groups of \citet{rentseek2026} contain two pairs with identical reward rules: a length penalty of 1.3 run twice, and an audit at rate 0.05 with fine 20 run twice. We merge each pair, which leaves 13 patches. They are seven random audits (rate 0.02 with fine 50; 0.05 with 20; 0.2 with 5; 0.05 with fines 2 and 6; a back-tax audit at rate 0.05; full gold), four length penalties ($\lambda \in \{1, 1.3, 3, 10\}$), a minimum over two proxies and a targeted audit. Some patch runs end at step 150, so the time course ends there (Table~\ref{tab:patches}).

\begin{table}[h]
\centering
\small
\setlength{\tabcolsep}{3pt}
\begin{tabular}{lccccc}
\toprule
 & All & Sign & Audits & Length & Demeaned \\
\midrule
Rent, step 50 & 0.91 & 0.85 & 0.79 & 0.80 & 0.73 \\
Rent, step 100 & 0.97 & 1.00 & 0.93 & 1.00 & 0.69 \\
Rent, step 150 & 0.99 & 1.00 & 1.00 & 1.00 & 0.79 \\
Gold, step 50 & 0.28 & 0.62 & $-$0.50 & 1.00 & 0.37 \\
Gold, step 100 & 0.40 & 0.54 & $-$0.32 & 1.00 & 0.47 \\
Gold, step 150 & 0.51 & 0.69 & $-$0.04 & 1.00 & 0.56 \\
\bottomrule
\end{tabular}
\caption{Retrospective forecast for 13 patches: Spearman correlation over all patches, share of correct signs, Spearman correlation within audits (7) and within length penalties (4), and after removal of type means.}
\label{tab:patches}
\end{table}

\section{Real RM Upgrades}\label{app:realrm}

From the branch points of three parents, RL continues with the original proxy in a control arm. In three further arms, the reward model changes to a model from \citet{rentseek2026}: Qwen models trained on 20,000 and 80,000 pairs, and a Llama model trained on 20,000 pairs. Each model has its own standardization constants, as in \citet{rentseek2026}. All other settings follow the factorial runs, for a total of 12 runs. We fixed the forecasts and four hypotheses before the runs.

\begin{table}[h]
\centering
\small
\setlength{\tabcolsep}{2.5pt}
\begin{tabular}{lrrrrrrr}
\toprule
Upgrade & F1a & F1b & F2 & F4 & $L$ & $\rho$ & $z_g$ \\
\midrule
Qwen, 20k & 0.51 & 0.25 & 0.12 & 0.07 & 0.82 & $-0.051$ & 0.053 \\
Qwen, 80k & 0.26 & 0.10 & 0.37 & 0.15 & 1.07 & $-0.065$ & 0.062 \\
Llama, 20k & 0.22 & 0.46 & 0.19 & 0.24 & 1.33 & $-0.033$ & 0.027 \\
\bottomrule
\end{tabular}
\caption{Pre-specified forecasts for real RM upgrades, mean over three parents. Supply in standardized units at the end window with $\kappa$ frozen from the factorial runs; rent and gold as $\Cov_B(\cdot, \dr)$.}
\label{tab:realrmpred}
\end{table}

\paragraph{Outcomes.} Table~\ref{tab:realrmout} lists the pre-specified hypotheses and their outcomes for parents P1--P3. The forecast of rent and gold reward holds in direction and in order at both windows. The observed changes at step 200 are larger than the first-order forecast, which we do not scale by $\kappa$ for rent. One sub-criterion fails: under the Llama upgrade, the policy adds 3.49 times the predicted number of headers, a ratio that is unstable because the predicted change is small (0.22 standard deviations). The KL that each upgrade accumulates beyond the control (0.023, 0.024 and 0.050) is ordered like the change in rent at step 200.

\begin{table}[h]
\centering
\small
\setlength{\tabcolsep}{3pt}
\resizebox{\columnwidth}{!}{%
\begin{tabular}{p{0.5\columnwidth}p{0.27\columnwidth}l}
\toprule
Hypothesis & Observed & Outcome \\
\midrule
U1: Stable supply directions are correct & 7/7 correct & Supported \\
U2: Frozen coefficients forecast supply sizes; covariance beats price & frozen $\kappa$: 0.66; refitted: 0.77 vs.\ 0.30 & Supported \\
U3: Early signs and ranking of rent and gold are correct & signs 9/9 (rent), 9/9 (gold); same order & Supported \\
U4a: No late reversal & same order at step 200 & Supported \\
U4b: Rise in headers is close to the forecast & observed / forecast: 1.32, 1.31, 3.49 & Not met (Llama) \\
\bottomrule
\end{tabular}}
\caption{Pre-specified hypotheses for real RM upgrades, parents P1--P3.}
\label{tab:realrmout}
\end{table}

\begin{table}[h]
\centering
\small
\setlength{\tabcolsep}{3pt}
\resizebox{\columnwidth}{!}{%
\begin{tabular}{lrrrrrr}
\toprule
 & \multicolumn{2}{c}{Forecast} & \multicolumn{2}{c}{Steps 20--30} & \multicolumn{2}{c}{Step 160--200} \\
\cmidrule(lr){2-3}\cmidrule(lr){4-5}\cmidrule(lr){6-7}
Upgrade & $\rho$ & $z_g$ & $\Delta\rho$ & $\Delta z_g$ & $\Delta\rho$ & $\Delta z_g$ \\
\midrule
Qwen, 20k & $-0.051$ & 0.053 & $-0.081$ & 0.053 & $-0.283$ & 0.211 \\
Qwen, 80k & $-0.065$ & 0.062 & $-0.103$ & 0.071 & $-0.380$ & 0.273 \\
Llama, 20k & $-0.033$ & 0.027 & $-0.066$ & 0.036 & $-0.136$ & 0.068 \\
\bottomrule
\end{tabular}}
\caption{Forecast (covariance at the branch point) and observed change in rent and gold reward relative to the control, mean over P1--P3.}
\label{tab:realrmrent}
\end{table}

\paragraph{Extension to six parents.} After the pre-specified analysis, we ran the same upgrades from the other three parents and report the six parents together. Over steps 20--30 and at step 200, all 18 forecast signs are correct for rent and all 18 for gold reward, and the six stable directions of supply are correct. With $\kappa$ frozen, the explained share of supply is 0.60 early and 0.74 at the end. Under the Llama upgrade, the policy again adds about three times the predicted number of headers (3.18).

\section{A Recipe for Comparing Reward Models before Training}\label{app:recipe}

\begin{enumerate}
\setlength\itemsep{0pt}
\item Sample $n \approx 16$ responses for each of a few hundred prompts from the current policy, with the decoding settings of RL.
\item Score every response with the current and the candidate reward model, and standardize each model with the constants that RL will use.
\item For every behavior $f$ of interest, compute the within-prompt covariance $\Cov_x(f, \dr)$ and average it over prompts. Divide by the within-prompt standard deviation of $f$ to compare behaviors.
\item Do the same for any available measure of quality, such as a stronger judge, to obtain the direction of the change in quality.
\item Read the signs as the forecast of direction. For sizes, fit one coefficient per behavior on a past edit of the same pipeline.
\item Track the KL divergence from the starting policy during RL. Once the divergence is large, the forecast of rent may no longer hold.
\item At a checkpoint, repeat steps 1--3 on samples of the current policy. If the covariance has grown, shrunk or changed sign, the forecast from the start no longer describes the incentive.
\end{enumerate}

\end{document}